\documentclass{article}
\usepackage[T1]{fontenc}
\usepackage{times}
\usepackage[margin=1in]{geometry}
\usepackage[round]{natbib}
\usepackage{amsmath}
\usepackage{graphicx}
\usepackage{xcolor}
\usepackage{hyperref}
\hypersetup{colorlinks=true,linkcolor=[rgb]{0,0.08,0.45},citecolor=[rgb]{0,0.08,0.45},urlcolor=[rgb]{0,0.08,0.45}}
\usepackage{url}
\usepackage{booktabs}
\usepackage{array}
\usepackage{microtype}

\definecolor{tagOffer}{RGB}{215,230,251}
\definecolor{tagInform}{RGB}{226,231,241}
\definecolor{tagAsk}{RGB}{250,244,206}
\definecolor{tagRefuse}{RGB}{251,214,218}
\definecolor{tagRequest}{RGB}{252,226,200}
\definecolor{tagAccept}{RGB}{211,239,214}
\definecolor{tagWireless}{RGB}{203,238,240}
\definecolor{tagMove}{RGB}{228,218,246}
\definecolor{tagTechvisit}{RGB}{214,228,240}
\definecolor{tagActual}{RGB}{253,230,191}
\definecolor{tagIntent}{RGB}{253,230,191}
\definecolor{tagOther}{RGB}{236,236,236}
\newcommand{\tagpill}[2]{{\setlength{\fboxsep}{2pt}\colorbox{#1}{\footnotesize\ttfamily #2}}}

\title{Clarify, Then Focus: Statement Normalization \\ for Conversation Analytics at Scale}

\author{Mikhail L. Arbuzov \quad Karan Dave \quad Evgeniya Dontsova \quad Yaodong Hu \\
Vincent Lao \quad Navita Jain \quad Sisong Bei \quad Dmitry Dimov \\[6pt]
\normalsize Independent researchers}

\date{}

\begin{document}
\maketitle

\begin{abstract}
Enterprise conversation analytics asks many questions of millions of interactions. Each question can require reconstructing what people mean and identifying which information matters, repeating costly interpretive work across the same transcripts. We propose a simple principle: clarify the text, then focus the reader. Statement normalization transforms dialogue into short, speaker-attributed statements with source references and semantic tags. The statements make meaning more explicit; the tags support selecting evidence for a particular question. Downstream models can use the full representation or a relevant subset, depending on what helps them make the decision. In an offer-suppression task on customer-service calls, normalization improves a supervised classifier without selection, while weaker prompted readers benefit from both normalization and selection. A small model can learn the normalization contract, while lightweight encoders handle tagging and downstream decisions. Sharing this preparation across questions supports an inference pipeline built entirely from small models, making analytics over millions of conversations substantially less expensive.
\end{abstract}
\section{Introduction}

A conversation happens once. Its meaning may be reconstructed many times.

An agent offers a Boost Mobile phone line for \$15.00 per month, mentioning its partnership with Dish. The customer wants to wait until after a move, asks for the TV service to be set up again, and plans to get a cell phone then. Which brands were discussed? What price was offered? Why was the offer deferred, and what does the customer intend to do next? Each question draws on a different part of the same exchange. Before answering, a model must first find these few turns within a conversation spanning thousands of tokens, then separate the offer, the customer's circumstances, and their future plans.

At enterprise scale, this interpretive work is repeated across large collections of long form transcripts. A direct prompted approach asks a language model to read the entire conversation for each task. A supervised classifier can make predictions more cheaply, but must learn its decision from transcripts containing disfluencies, implicit references, and evidence distributed across turns. The input representation therefore affects what the downstream model must do.

This motivates a different division of work: \textbf{clarify the conversation once, then focus the reader for each decision.} Normalization decomposes dialogue into short, simple, attributed statements with source references and canonical entity names. Semantic tags make the statements selectable. The representation is constructed without a downstream question and stored for subsequent analysis.

This division of work supports a pipeline built entirely from small models. On our offer-suppression pilot, a 0.6B normalizer and a task-trained encoder achieve 0.839 F1, compared with 0.857 for Haiku on full teacher-normalized input. At the reported rates, the pipeline has approximately 100-fold lower projected serving cost than Haiku's reader alone at twenty separately scored questions per conversation (Section 6).

\begin{figure}[t]
\centering
\includegraphics[width=\linewidth]{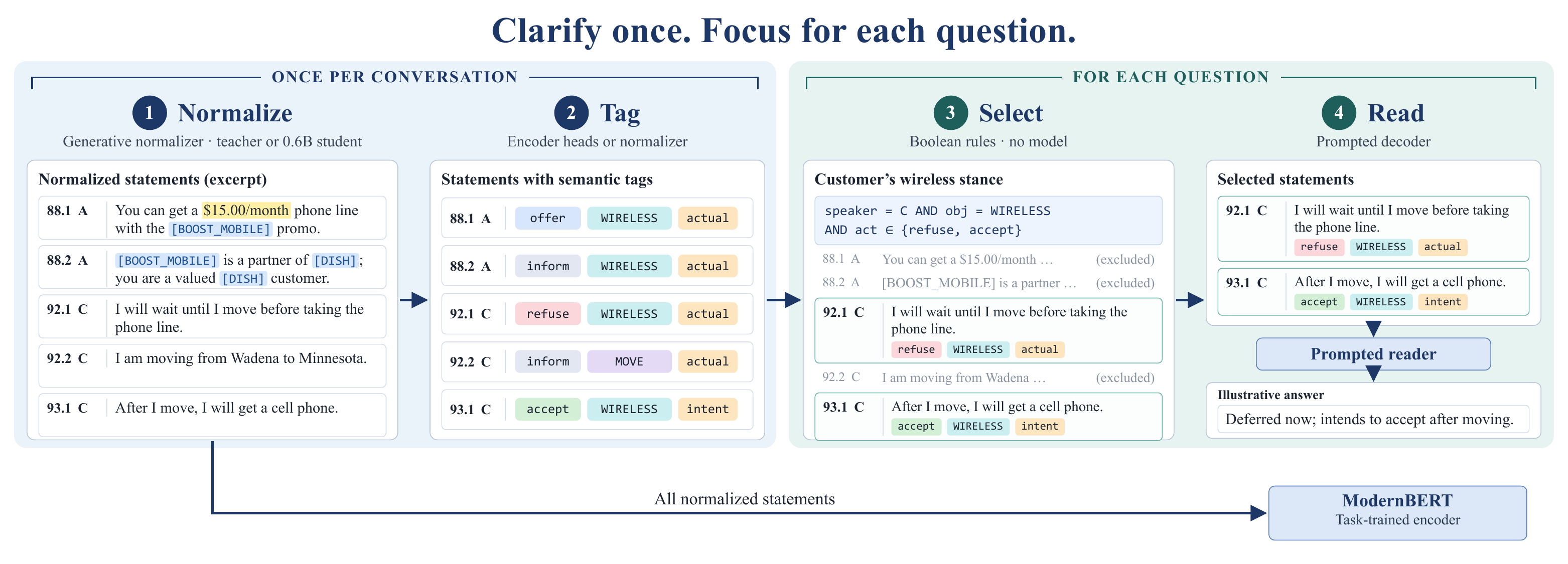}
\caption{\textbf{Clarify once, focus for each question.} Normalization and tagging run once per conversation and are stored. For each question, a Boolean predicate selects statements for a prompted reader, or a task-trained encoder reads all statements. The panels use an excerpt of Example 1.}
\label{fig:figure1_pipeline}
\end{figure}

\noindent\begin{minipage}{\linewidth}
\textbf{Example 1. One exchange, nine tagged statements.} The statements span six source turns (88--93). \texttt{A} denotes the agent and \texttt{C} the customer. Each line ends with tags in the order \texttt{speech act $\cdot$ subject $\cdot$ qualification}.
\begin{quote}\small
\begin{minipage}{\linewidth}\raggedright\setlength{\parindent}{0pt}
\textbf{88.1 A} You can get a \$15.00/month phone line with the [BOOST\_MOBILE] promo.\enspace\allowbreak\mbox{\tagpill{tagOffer}{offer}\,\tagpill{tagWireless}{WIRELESS}\,\tagpill{tagActual}{actual}}\par\smallskip
\textbf{88.2 A} [BOOST\_MOBILE] is a partner of [DISH]; you are a valued [DISH] customer.\enspace\allowbreak\mbox{\tagpill{tagInform}{inform}\,\tagpill{tagWireless}{WIRELESS}\,\tagpill{tagActual}{actual}}\par\smallskip
\textbf{89.1 A} We are giving you a \$15.00 phone line.\enspace\allowbreak\mbox{\tagpill{tagOffer}{offer}\,\tagpill{tagWireless}{WIRELESS}\,\tagpill{tagActual}{actual}}\par\smallskip
\textbf{90.1 C} Is the phone line a home phone or a cell phone?\enspace\allowbreak\mbox{\tagpill{tagAsk}{ask}\,\tagpill{tagWireless}{WIRELESS}\,\tagpill{tagActual}{actual}}\par\smallskip
\textbf{91.1 A} The phone line can be a home phone or a cell phone.\enspace\allowbreak\mbox{\tagpill{tagInform}{inform}\,\tagpill{tagWireless}{WIRELESS}\,\tagpill{tagActual}{actual}}\par\smallskip
\textbf{92.1 C} I will wait until I move before taking the phone line.\enspace\allowbreak\mbox{\tagpill{tagRefuse}{refuse}\,\tagpill{tagWireless}{WIRELESS}\,\tagpill{tagActual}{actual}}\par\smallskip
\textbf{92.2 C} I am moving from Wadena to Minnesota.\enspace\allowbreak\mbox{\tagpill{tagInform}{inform}\,\tagpill{tagMove}{MOVE}\,\tagpill{tagActual}{actual}}\par\smallskip
\textbf{92.3 C} You will have to come set the TV back up after I move.\enspace\allowbreak\mbox{\tagpill{tagRequest}{request}\,\tagpill{tagTechvisit}{TECH\_VISIT}\,\tagpill{tagActual}{actual}}\par\smallskip
\textbf{93.1 C} After I move, I will get a cell phone.\enspace\allowbreak\mbox{\tagpill{tagAccept}{accept}\,\tagpill{tagWireless}{WIRELESS}\,\tagpill{tagIntent}{intent}}
\end{minipage}
\end{quote}
\end{minipage}

These units are ``greppable.'' A wireless tag retrieves the offer and the customer's responses even when they do not name either brand. Combining subject, speaker, and speech act isolates the agent's offers; other predicates expose the move, TV setup, or future intent. A downstream model can read all statements or a selected subset, according to what helps its decision.

Normalization may appear to require a large model. Its output, however, follows a stable contract: separate claims, preserve speakers and qualifications, and resolve references into explicit statements. We treat this as a constrained semantic mapping that a small model can learn from a teacher. This makes both construction and consumption of the representation candidates for inexpensive inference.

We evaluate the design on an offer-suppression task in customer-service calls. Our contributions connect the representation to its downstream use:

\begin{enumerate}

\item \textbf{Statement normalization with canonical entities.} We develop attributed units intended to capture individual claims, preserving source references and canonical names for known entities. These units make conversational information easier for downstream models to read and for rules to address.

\item \textbf{Programmatic selection through semantic tags.} Speech-act, subject, and qualification tags combine with speaker and entity predicates to expose fine-grained evidence slices. New combinations can be selected from the stored representation without another generative pass.

\item \textbf{Utility for encoder and decoder readers.} We compare raw and normalized encoder inputs and three input representations across nine prompted readers. The encoder benefits without selection; several weaker prompted readers benefit from both normalization and further selection.

\item \textbf{Small models for preparation and prediction.} We substitute a 0.6B normalizer under an unchanged production classifier and measure the resulting quality trade-off. We connect this serving configuration to a cost model that separates one-time preparation from recurring decisions. Three tagging heads demonstrate shared encoder computation; downstream reuse across questions is projected.

\end{enumerate}

\section{Related work}

\textbf{Explicit statements and rewriting.} Decontextualization aims to preserve a sentence's meaning while making it interpretable outside its original context \citep{choi-etal-2021-decontextualization}. Discourse-aware simplification separates propositions while retaining their relationships \citep{niklaus-etal-2022-shallow}; MinIE represents polarity, modality, attribution, and quantities alongside compact relational tuples \citep{gashteovski-etal-2017-minie}. These relationships and qualifications matter when decomposition could detach a claim from its conditions. Dense X Retrieval uses self-contained propositions as retrieval units and trains their producer from large-model examples \citep{chen-etal-2024-dense}. It establishes precedents for both decomposition and a learned producer. Our evaluation concerns attributed conversation statements with canonical entities and controlled tags as inputs to call-level decisions.

Rewriting also provides an interface to downstream models. \citet{su-etal-2019-improving} recover omitted and referential content in contextual utterances and report gains in dialogue systems. This targets a current request; we construct an inventory of statements for subsequent questions. Telegraph English rewrites text into compact, addressable units and evaluates smaller readers, but its QA experiments do not test its proposed tag-based selection mechanism \citep{arbuzov2026telegraphenglish}. A companion study on multi-hop question answering finds that such readable symbolic re-expression preserves entity content more densely than matched-budget deletion, truncation, or a coherent prose summary \citep{bei2026contextcompression}. We explicitly compare the full representation with selected statements and include an encoder that benefits without selection.

\textbf{Metadata and selective access.} Dialogue-act annotation distinguishes communicative functions and qualifiers on functional segments \citep{bunt-etal-2017-dialogue}. Metadata can also support retrieval: A-MEM enriches interactions with descriptions, keywords, and tags, embeds the records, and retrieves by semantic similarity \citep{xu2025amem}. SimpleMem combines selective retention, write-time synthesis, and model-planned retrieval over semantic, lexical, and symbolic indexes for long-term agent memory \citep{liu2026simplemem}. Our controlled tags serve a simpler access interface: fixed Boolean predicates over statements, speakers, and entities. Their value is evaluated through the behavior of the downstream reader.

\textbf{Compression and reuse.} RECOMP uses query-focused compression in its QA setting \citep{xu2024recomp}; AttentionRAG selects original sentences using query-conditioned attention \citep{fang2025attentionrag}. Query-independent alternatives include LLMLingua-2's learned word selection \citep{pan-etal-2024-llmlingua} and ReadAgent's reusable gists with question-dependent passage lookup \citep{pmlr-v235-lee24c}. Offline extraction for later QA \citep{fleischman-etal-2003-offline} and EVAPORATE's structured views and generated extraction functions \citep{arora2023evaporate}, and precomputed retrieval indexes such as GraphRAG and RAPTOR \citep{edge2024graphrag,sarthi2024raptor}, likewise move work ahead of individual questions. Query independence alone is therefore not our contribution. We investigate the combination of readable statements, controlled selection, encoder and decoder utility, and a small learned producer.

\textbf{Reusable decision readers.} Jev accepts textual state and question-defined output schemas to produce typed decisions \citep{almeida2026jev}. Laya provides an open-weight Jev-compatible interface, with a ModernBERT backbone and a learned decision head \citep{convaiinnovations2026laya}. These readers make it possible to specify new decisions without fitting a separate head for each question. Our representation supplies their evidence; our Laya experiment tests transfer to a domain-specific policy. Appendix A develops these comparisons and additional related work.

\section{Method}

\subsection{Constructing attributed statements}

A conversation is hard to read because a single claim is rarely contained in a single turn. It may be split across turns by the speech recognizer, buried under disfluencies, or carried by a pronoun whose referent was established minutes earlier. Normalization rewrites each bounded window of turns into short, self-contained, first-person statements, so that a claim can be read on its own without replaying the call around it. Filler, backchannel, and greetings are dropped; a turn that runs several claims together becomes one statement each, and a sentence the recognizer split across turns is rejoined under the turn where it begins. Every number, price, date, and name is restated exactly rather than described, redaction placeholders such as \texttt{[CREDIT\_DEBIT\_NUMBER]} that arrive already typed from an upstream pass are copied character for character, and known brand variants are mapped to canonical tokens such as \texttt{[BOOST\_MOBILE]} and \texttt{[DISH]} by a deterministic gazetteer. Every statement keeps the global index of the turn it came from: in Example 1, the integer identifies that source turn and the suffix distinguishes statements within it. The discipline throughout is to resolve and restate, never to invent or expand --- a pronoun is replaced only by an antecedent the window already supplies, a short utterance is not inflated into an intent the surrounding turns do not establish, and each speaker's stance is kept, so a refusal stays a refusal.

A few turns from a synthetic call show the effect; Appendix B.5 works the whole exchange end to end. Three raw recognizer turns,

\begin{itemize}

\item \textbf{[5] C:} I wanted to add the, uh, the sports package, for the college football this weekend.

\item \textbf{[10] A:} sure, so it's uh \$13.00 a month, but I can do 50\% off, so that's \$6.50.

\item \textbf{[19] C:} no, I'm good for now.

\end{itemize}

become self-contained statements, shown here with the tags of Example 1:

\begin{quote}\small
\begin{minipage}{\linewidth}\raggedright\setlength{\parindent}{0pt}
\textbf{5.1 C} I want to add the sports package for the college football games this weekend.\enspace\allowbreak\mbox{\tagpill{tagRequest}{request}\,\tagpill{tagOther}{SPORTS\_PACKAGE}\,\tagpill{tagActual}{actual}}\par\smallskip
\textbf{10.1 A} The [MULTI\_SPORT\_PACK] is \$13.00 a month, but I can add it at 50\% off, which is \$6.50.\enspace\allowbreak\mbox{\tagpill{tagOffer}{offer}\,\tagpill{tagOther}{SPORTS\_PACKAGE}\,\tagpill{tagActual}{actual}}\par\smallskip
\textbf{19.1 C} I am not interested in [BOOST\_MOBILE] right now.\enspace\allowbreak\mbox{\tagpill{tagRefuse}{refuse}\,\tagpill{tagWireless}{WIRELESS}\,\tagpill{tagActual}{actual}}
\end{minipage}
\end{quote}

Filler is dropped, the price is restated exactly, the agent's brand is canonicalized to \texttt{[MULTI\_SPORT\_PACK]}, and ``no, I'm good for now'' is resolved into a refusal of the \texttt{[BOOST\_MOBILE]} service named a turn earlier --- kept as a refusal rather than softened into a deferral. A reader can act on any one of these statements without replaying the call around it. Because the same fixed rules apply to every window, normalization is a constrained mapping a small distilled model can learn (Section 3.3). The normalizer itself performs no typing, entity extraction, or numbering, which are separate downstream stages.

Each statement then carries three tags drawn from fixed vocabularies. The speech-act tag (\texttt{act}, 10 values) records what the utterance does, such as \texttt{offer}, \texttt{refuse}, or \texttt{accept}. The subject tag (\texttt{obj}, 33 values) records the business topic, such as \texttt{WIRELESS}, \texttt{MOVE}, or \texttt{TECH\_VISIT}. The qualification tag (\texttt{mod}, 6 values) records how the claim is held, for example as an \texttt{actual} event or a future \texttt{intent}. Together the three families describe up to 10 $\times$ 33 $\times$ 6 = 1,980 statement types. This counts only the controlled tags; the canonical entity tokens --- nine closed keys such as \texttt{product} (\texttt{[BOOST\_MOBILE]}) and \texttt{competitor} (\texttt{[VERIZON]}), each with an open vocabulary of values --- add a further selection axis through the literal-match predicates in Appendix B.1. The tag vocabularies are fixed in advance by the normalization contract rather than learned from data.

Tags label normalized statements rather than raw turns, and this ordering matters. A raw turn often mixes several claims: customer turn 92 in Example 1 contains a deferral, a move, and a service request. A single tag on that turn would have to describe all three at once. Normalization first separates the claims, so that each tag describes one of them.

Tags can come from two producers. The generator can append them while writing each statement, or a separate encoder can assign them afterwards, using one shared trunk with a classification head per family. The two routes differ in cost and in accuracy, so the chosen producer must be counted in the construction cost and its output checked against the contract. In both cases, construction receives no downstream question. The schema is still domain-specific: its subject vocabulary names the recurring topics of this business, which is what makes later selection possible.

\subsection{Selecting evidence and predicting}

For transcript \(X\), let \(Z = N(X)\) be the normalized representation. A task-specific selector \(S_q\) supplies evidence to reader \(H_q\):

\[
\hat{y}_q = H_q\bigl(S_q(Z)\bigr).
\]

Here \(\hat{y}_q\) is the reader's predicted decision for question \(q\). \(S_q\) either retains all statements or applies Boolean predicates over tags, speaker, and canonical-entity matches. Selection requires no further model call once the representation exists. In Example 1, \texttt{speaker = A AND act = offer AND obj = WIRELESS} retrieves the two offers (88.1 and 89.1). A stance decision may need both the deferral in 92.1 and future acceptance in 93.1; selecting only refusals would omit relevant evidence. Appendix B.1 gives exact slices and discusses context dependencies.

Two kinds of reader consume the representation. The supervised reader is a ModernBERT encoder with a task-specific classification head \citep{warner-etal-2025-smarter}, trained on task labels. The prompted readers are decoder models given a fixed task rubric and no task-specific training. Either kind can be fed the full normalized call or the subset chosen by a subject-match selector --- the STOP selector for the offer-suppression task of Section 4, which keeps statements whose subject tag matches the offered product; which input serves a given reader is decided by task performance, not by a fixed rule. In the ablations below, the encoder is compared on raw and full normalized input, and each prompted reader additionally receives the selected statements.

Comparing inputs within each reader separates the effect of normalization from the effect of selection. Selection is an input variant a reader uses when it helps, not a prerequisite for benefiting from normalization. Section 5 reports which readers gain from it.

\subsection{Learning a small producer}

Llama 4 Maverick supplies normalized statements and task labels for the training corpus. The statements supervise smaller normalizers through teacher-output training \citep{kim-rush-2016-sequence}; the task labels separately supervise classifiers. The normalizer learns the representation, while the classifier learns a decision over it.

We distill two student normalizers, of 0.6B and 1.7B parameters, and train each size with three seeds. Several seeds are needed because, at this corpus size, a single run cannot separate a real difference between sizes from training randomness.

The serving comparison asks the deployment question directly. The production classifier is trained once on teacher-normalized text and then held fixed. At test time, its input is switched from teacher output to student output without retraining. This mirrors practice, where the normalizer may be replaced without retraining every downstream classifier. One rendering detail is controlled. The teacher emits per-statement turn numbers during normalization and the student does not, so both inputs receive identically synthesized numbering. Re-rendering the teacher input this way reproduces its original score within a pre-declared tolerance, so a remaining gap reflects the normalizer rather than formatting.

\section{Experimental setup}

\textbf{Task and data.} STOP is a call-level operational label indicating whether further offers of the pitched product to this customer should be suppressed. A complaint, product mention, or refusal tag alone does not define the decision. Evaluation measures agreement with that label.

The corpus contains 194,042 English customer-service calls with teacher normalization and teacher task labels. The production classifier is trained on 151,914 calls. The raw-versus-normalized ablation uses 16,050 held-out calls that fit the context limit in both representations, covering 96.6\% of distinct held-out call IDs. These silver-label results measure agreement with the teacher that also constructs the normalized input.

Independent human evaluation starts from 80 calls annotated by two people. All reported human scores use the fixed 66-call consensus subset, including 17 STOP positives. Consensus filtering limits the population evaluated; one positive call changes recall by approximately 0.059. Small differences do not establish superiority or equivalence.

The human set is small by necessity, and this reflects the setting. The STOP label encodes a business policy, namely when the company should stop offering a product. It must therefore come from domain experts who own that policy, and general crowd annotators cannot supply it. Expert time is scarce, so enterprise deployments typically combine a large model-labeled corpus with a small expert-labeled set for validation. We design the analysis around that regime.

\textbf{Comparisons.} The encoder ablation compares classifiers of the same architecture trained on raw and normalized inputs. Its normalized-input classifier is distinct from the production classifier used for normalizer substitution. Nine prompted readers each receive raw, full normalized, and selected inputs under the same task rubric and denominator. This separates raw-to-full and full-to-selected changes within each reader. Finally, the unchanged production classifier receives teacher or student normalization. The separate exploratory zero-shot protocol is reported in Appendix D.

\section{Results}

The results follow the proposed decomposition: clarify the input for an encoder, focus it for prompted readers, and replace the large producer with a small one. We then test whether a reusable decision reader can apply the policy without task-specific training. Human scores share the 66-call denominator; the larger silver-label analysis is identified separately.

\subsection{Clarifying the input helps the supervised encoder}

\textbf{Table 1. Encoder ablation on 66 human-consensus calls, including 17 positives.} These classifiers differ from the production classifier in Table 3.

\begin{center}
\small
\setlength{\tabcolsep}{3pt}
\begin{tabular}{lll}
\toprule
\textbf{Input representation} & \textbf{STOP F1} & \textbf{ROC-AUC} \\
\midrule
Raw transcript & 0.7879 & 0.9304 \\
Normalized statements & 0.8235 & 0.9616 \\
\bottomrule
\end{tabular}
\end{center}

Normalization increases both reported metrics without selecting evidence. The larger silver-label comparison is consistent in direction: precision increases by 9.8, 11.9, 15.6, and 15.8 percentage points at the reported recall settings of 0.60, 0.70, 0.80, and 0.90, respectively (Appendix B.2). Those gains measure improved teacher agreement; the human pilot supplies independent but less precise evidence.

\subsection{Focusing the input further helps several prompted readers}

\textbf{Table 2. STOP F1 for nine prompted readers on the same 66 calls.} \texttt{Raw} is the raw transcript, \texttt{Full norm.} the complete normalized call, and \texttt{Selected norm.} the subset of normalized statements chosen by the subject-match selector; the final column is the F1 difference between selected and raw input. Scores are from the 66-call pilot, so per-reader differences are small and mixed in direction.

\begin{center}
\small
\setlength{\tabcolsep}{3pt}
\begin{tabular}{lllll}
\toprule
\textbf{Reader} & \textbf{Raw} & \textbf{Full norm.} & \textbf{Selected norm.} & \textbf{\(\Delta\)(sel, raw)} \\
\midrule
glm-4.7-flash & 0.4545 & 0.5833 & 0.7857 & +0.3312 \\
nova-micro & 0.4516 & 0.5405 & 0.7692 & +0.3176 \\
nemotron-nano-12b & 0.3810 & 0.5185 & 0.6364 & +0.2554 \\
llama4-scout-17b & 0.5833 & 0.5833 & 0.6897 & +0.1064 \\
nova-lite & 0.6222 & 0.5957 & 0.7317 & +0.1095 \\
llama4-maverick-17b & 0.7568 & 0.8125 & 0.7895 & +0.0327 \\
claude-haiku-4.5 & 0.8235 & 0.8571 & 0.8235 & 0.0000 \\
llama-3.3-70b & 0.8000 & 0.7333 & 0.7500 & -0.0500 \\
qwen3-32b & 0.7000 & 0.6250 & 0.5614 & -0.1386 \\
\bottomrule
\end{tabular}
\end{center}

The median paired F1 change is +0.0336 from raw to full input and +0.1064 from full to selected input. For several lower-scoring readers, both stages help: GLM improves from 0.4545 to 0.5833 to 0.7857, and nova-micro from 0.4516 to 0.5405 to 0.7692. Selection adds value beyond full normalization in these configurations.

The effect depends on the reader. Haiku performs best on the full normalized call; Qwen degrades across both transformations. Thus, clearer text and a narrower input are separate interventions whose usefulness must be checked for the chosen reader. Because selection operates on normalized statements, this comparison does not establish the benefit of rewriting over an evidence-matched selection of original turns. Appendix B.3 reports the exploratory cross-reader associations.

\subsection{A small producer supports competitive task quality}

Replacing teacher normalization with a 0.6B student under an unchanged production encoder yields a serving pipeline with no large-model inference call.

\textbf{Table 3. Changing the normalizer under the same production classifier.} Metrics use the 66-call human-consensus pilot. The classifier differs from the ablation in Table 1.

\begin{center}
\small
\setlength{\tabcolsep}{3pt}
\begin{tabular}{lllll}
\toprule
\textbf{Input producer} & \textbf{Precision} & \textbf{Recall} & \textbf{STOP F1} & \textbf{ROC-AUC} \\
\midrule
Teacher normalizer & 0.8750 & 0.8235 & 0.8485 & 0.9676 \\
0.6B student normalizer & 0.9286 & 0.7647 & 0.8387 & 0.9556 \\
\bottomrule
\end{tabular}
\end{center}

The student pipeline reaches 0.8387 F1, exceeding the best reported configurations of Llama 3.3 70B (0.8000) and Maverick (0.8125), as well as Haiku on raw or selected input (0.8235). Haiku on full teacher-normalized input achieves 0.8571, the highest prompted F1 in the reported sweep. We therefore use that configuration as the quality-relevant cost reference in Section 6.

Student substitution increases precision and reduces recall. The F1 change is -0.0098, with a reported bootstrap interval of [-0.0544, +0.0185], which includes both zero and practically relevant loss. At the reported 0.90 recall setting, precision is 0.5484 with student input versus 0.8421 with teacher input. This high-recall gap remains material despite the close headline F1; the pilot's coarse recall steps require the operating points to be interpreted through their achieved recall and thresholds.

Student output contains approximately 11\% more statements and 9\% more tokens per call, without establishing the cause of the quality change. Across three seeds per size, within-size variation exceeds the small 0.6B-versus-1.7B difference. The experiments therefore support a working small-model substitution, with quality assessed at the intended operating point.

\subsection{Zero-shot transfer of a reusable decision reader}

A reusable reader could answer new questions from their definitions, reducing the need to train and maintain a separate classifier for each task. We evaluate the released Laya typed-decisions checkpoint \citep{convaiinnovations2026layatyped}, an open-weight Jev-compatible reader, on selected normalized statements without STOP-specific training. Across four compact task definitions, ROC-AUC ranges from 0.607 to 0.733, compared with 0.9616 for the task-trained encoder on the full normalized call. The released checkpoint does not recover the supervised reader's quality on this policy. Training, input scope, and definition length differ; Appendix D reports the protocol and all four variants.

\section{Quality and cost at scale}

At twenty questions per conversation, the reported rates give a projected serving cost of about \$398 per million calls for the student pipeline, including normalization, versus \$38,319 for Haiku's reader alone. This is a 96.2-fold difference, or approximately 99\% lower cost. The projection combines inexpensive normalization with cheap recurring encoder decisions.

Let \(B\) be the shared cost of normalization and required metadata, \(c_q\) the inference cost for question \(q\), and \(n\) the number of distinct questions. The serving cost is

\[
C(n) = B + \sum_{q=1}^{n} c_q.
\]

The reported rates per million calls are a student normalization cost \(B_S = \$264.85\), \$6.68 per encoder pass, and \$1,915.95 per question for Haiku reading the full normalized call. These are conditional estimates on the reported offline batch/list-price basis, not invoice-verified spend. The student rate is paired with its measured F1 of 0.8387. Haiku's 0.8571 uses teacher-normalized input, whose separate construction cost \(B_T\) remains unresolved:

\[
C_{\mathrm{student}}(n) = B_S + 6.68n, \qquad C_{\mathrm{Haiku}}(n) = B_T + 1{,}915.95n.
\]

\textbf{Table 4. Observed STOP F1 and projected serving cost per million calls (dollars).} Student costs include normalization; Haiku costs exclude \(B_T\). \(n > 1\) projects separately scored questions at the stated rates, not measured multi-task quality. Costs are rounded.

\begin{center}
\small
\setlength{\tabcolsep}{3pt}
\begin{tabular}{p{0.626\linewidth}p{0.086\linewidth}p{0.061\linewidth}p{0.074\linewidth}p{0.074\linewidth}}
\toprule
\textbf{Serving configuration} & \textbf{STOP F1} & \textbf{n = 1} & \textbf{n = 20} & \textbf{n = 50} \\
\midrule
0.6B normalizer + encoder & 0.8387 & 272 & 398 & 599 \\
Haiku on full teacher-normalized input, reader only & 0.8571 & 1,916 & 38,319 & 95,798 \\
\bottomrule
\end{tabular}
\end{center}

The ratio is 7.1 for one question and 160.0 for fifty questions. Adding Haiku's teacher-normalization cost, \(B_T\), increases these ratios. The measured quality comparison is 0.8387 versus 0.8571 STOP F1; the multi-question workloads project separate task passes at the stated rates.

The comparison concerns serving under separate task passes. It excludes labeling, training, calibration, and maintenance. Multi-question prompts, provider caching, shared encoder heads, summaries, and retrieval indexes are alternative reuse strategies that need their own quality and cost measurements. Rates must also cover the evaluated input lengths and any required metadata production. Appendix C gives the detailed projections, lower-priced baselines, and accounting assumptions. The result supports a conditional cost advantage at nearby observed F1, not universal cost dominance.

\section{Discussion and limitations}

\textbf{Evidence and scope.} The human pilot covers one task in one English customer-service domain and cannot establish narrow performance differences or transfer to new tasks. Consensus filtering can exclude difficult cases. The larger evaluation measures agreement with the same teacher that creates the representations. Reuse is already exercised at the tagging stage, where three classification heads share one encoder trunk over the same stored statements. Adding further downstream questions follows the same pattern: a new selector and head over an unchanged representation. We have not yet measured end-to-end quality for several downstream questions at once.

\textbf{What the intervention changes.} Rewriting, canonicalization, tagging, and selection are coupled by design. Selection is only as reliable as the tags, and tags are only as reliable as the units they label. Tagging raw turns would attach one label to several mixed claims, so normalization is what makes tag-based selection possible. The coupling still leaves each stage's separate contribution open. An evidence-matched original-turn control, surface cleanup, summary, and raw-text retrieval baselines are needed to isolate their contributions. Normalization can lose qualifications and selection can discard necessary evidence; source fidelity and tagging correctness require independent evaluation. Reuse is limited to information the representation retains.

\textbf{Can one decision model serve many questions?} Task-trained heads currently provide the stronger STOP results, while reusable decision readers could reduce the labeling, calibration, and maintenance work that recurs with each new question. The stored representation supports either route. A reusable reader must recover the decision rule from its definition as well as interpret the evidence. Further comparisons should measure the benefit of normalization and selection for reusable readers and weigh any quality loss against the reduction in per-question setup.

\section{Conclusion}

Conversation analytics repeats the same interpretive work every time a new question is asked of an old transcript. We propose doing that work once. Statement normalization turns each conversation into short, attributed, tagged statements, stored for later questions. For each question, a downstream reader then consumes either all statements or a slice chosen by a Boolean predicate.

On the offer-suppression task, normalization improves the supervised encoder without selection, while several weaker prompted readers benefit from both clearer statements and a narrower evidence set. A 0.6B normalizer supplies a fixed production encoder at 0.8387 F1, close to the best reported prompted configuration's 0.8571. At the reported rates, this pipeline has approximately 100-fold lower projected serving cost at twenty questions per conversation. Laya's zero-shot results identify a remaining challenge: extending the strong task-trained results to a reader that accepts new decision definitions without a separate trained head.

The human evidence covers one business policy and a small expert-labeled cohort; multi-question savings are projected. Within this setting, the results support a practical division of work: make meaning explicit once, select evidence when it helps, and use small models for recurring decisions. Evaluating more questions over the stored representation will establish how far that reuse extends.

\section{Ethics statement}

The study analyzes enterprise customer-service calls, and the evaluated label concerns suppression of further product offers. Errors can change which customers continue to receive offers. Aggregate agreement alone does not establish that two systems affect the same individuals, so operational use requires calibration and review at the intended decision threshold.

Access to the underlying transcripts is restricted by the deployment's data agreement. Brand canonicalization and speaker attribution are representation operations; they should not be interpreted as guarantees of anonymization. The example is an attributed agent--customer exchange. Its statements record what the speakers said, including the agent's offer and the customer's plans, without independently verifying those claims.

\section{Reproducibility statement}

The main tables distinguish encoder ablation, prompted-reader inputs, and fixed-classifier normalizer substitution. Appendices B--D report supplementary analyses, cost assumptions, and the exploratory Laya protocol. The private transcript corpus is not publicly released, limiting exact replication. Reimplementation on another corpus would test transfer rather than reproduce these particular scores.

\section{AI-use statement}

AI assistance was used to revise the paper's narrative, review methodological choices and interpretations, check consistency among reported quantities, and propose further analyses. Language models were also used within the experimental pipeline to construct representations and supervision and to make prompted predictions. The authors are responsible for verifying the final text, results, citations, and disclosures against the underlying research records.

\appendix

\section{Extended related work}

Our pipeline makes three commitments, and each one meets a separate line of prior work. The first concerns the unit into which a conversation is rewritten: short statements that carry one claim, a speaker, a source reference, and a qualification. The second concerns how those units reach a reader: either all at once or through a Boolean predicate over controlled tags. The third concerns when interpretation is paid for and by which model: once per conversation, before any question is known, by a small distilled producer. Much of each commitment has precedent. This appendix follows those precedents in turn, noting what each line of work established, the setting it was built for, and the part of the problem it leaves open. The final subsection states which combination we believe is new.

\subsection{From sentences to self-contained units}

The idea that text becomes easier to use once it is broken into small, self-contained pieces has a long history. Split and Rephrase framed the task directly: rewrite a complex sentence as a meaning-preserving sequence of shorter sentences, supported by a benchmark of over a million complex-to-simple examples \citep{narayan2017split}. Decontextualization added a second requirement. A sentence taken from a Wikipedia article should be rewritten so that it can be interpreted without its surrounding context, while keeping its meaning \citep{choi-etal-2021-decontextualization}. Splitting alone can lose the relations between the pieces, and discourse-aware simplification addresses this by producing a hierarchy of core and context sentences linked by rhetorical relations \citep{niklaus-etal-2022-shallow}. Open information extraction reached a similar concern from the opposite direction. MinIE shortens relational tuples but moves polarity, modality, attribution, and quantities into explicit annotations, so that minimizing a fact does not silently change what it asserts \citep{gashteovski-etal-2017-minie}.

Recent work combines these requirements in the atomic proposition. Dense X Retrieval defines a proposition as a distinct, minimal, and self-contained piece of meaning, and indexes English Wikipedia as roughly 257 million of them \citep{chen-etal-2024-dense}. Its Propositionizer is trained by distillation: GPT-4 decomposes 42k passages, and a Flan-T5-large model is fine-tuned on the result. FActScore decomposes generated text into atomic facts in order to measure factual precision \citep{min2023factscore}.

This line of work supplies the design principles our statements inherit, and Dense X shows that a large model's decomposition can be taught to a small one. The text it was developed on, however, is written, monologic, and largely declarative. A conversation adds problems these methods were not built for. Every claim has a speaker, and the same words mean different things from an agent and from a customer. Positions change within a call, so that a deferral in one turn and a conditional acceptance in the next are both part of the customer's stance. Many claims are commitments or intentions rather than facts, and their conditions, such as ``after I move,'' carry the business meaning. Our statements therefore add speaker attribution, links back to source turns, and an explicit qualification tag to the proposition principles. They are judged by whether downstream decisions improve, rather than by retrieval granularity or factual precision.

\subsection{Rewriting dialogue for the next component}

Dialogue research arrived at rewriting through a different need. In a preliminary study of 2,000 Chinese multi-turn conversations, \citet{su-etal-2019-improving} found coreference or omission in more than 70\% of utterances. They collected 20k annotated dialogues, trained a pointer-based Transformer to rewrite each utterance so that it recovers referenced and omitted content, and integrated the rewriter into two online chatbots, where it improved intention detection and user engagement. \citet{rastogi-etal-2019-scaling} treated reference resolution in multi-domain spoken dialogue as context-aware query reformulation, so that downstream language-understanding components receive a self-contained request.

These systems show that a rewritten conversation is a better interface for downstream models than the raw one, which is also our premise. Their rewriting is prospective and local, however. It runs online, one turn at a time, and serves a single question: what does the user want now, so that the system can act next. Our normalization is retrospective and global. It runs after the call has ended, over the whole conversation, and produces an inventory of every attributed claim for questions that no one has asked yet.

Simplification has also been tested as preprocessing for specific downstream tasks. Simple-SQuAD rewrites SQuAD contexts with a style-transfer simplifier, locates each answer span again in the simplified text, and reports gains of up to 2.04 exact-match points \citep{dadu-etal-2021-text}. DisSim-FinBERT applies discourse simplification to Federal Open Market Committee minutes, improving aspect selection and sentiment prediction \citep{kim2025dissimfinbert}. Both results confirm that simpler units can help a reader, but in each case the transformation is built and validated around one known task. Simple-SQuAD, in particular, repairs answer positions after simplification, which is possible only because the answers are known in advance. A representation for analytics has to preserve evidence for answers that do not yet exist.

\subsection{Structure that makes text addressable}

Labeling what an utterance does is an established part of dialogue analysis. The ISO 24617-2 standard annotates functional segments with communicative functions across several dimensions, together with qualifiers for properties such as certainty, conditionality, and sentiment \citep{bunt-etal-2017-dialogue}. Our speech-act and qualification tags echo this design. The standard, however, is a general scheme for describing dialogue. Our tags form a small, application-specific vocabulary whose subject family names business topics, and their purpose is operational: they exist so that a rule can select evidence.

The closest prior proposal is Telegraph English \citep{arbuzov2026telegraphenglish}. It rewrites text, without reference to any question, into atomic fact lines that use about forty logical and relational symbols. It tags lines for temporal state, modality, semantic role, and scope, and argues that each line becomes an independently addressable unit that a query can retrieve together with its scope. Its evaluation measures the rewrite as static compression, over 4,081 LongBench-v2 questions and five OpenAI models, at compression ratios matched against LLMLingua-2. At about half the original tokens, it preserves 99.1\% of key-fact accuracy with GPT-4.1, and its advantage grows for smaller models, reaching 11 points on fine-detail questions. Its authors state that dynamic context management is not yet benchmarked, and they describe compress-once reuse as an architectural argument rather than an empirical result.

Our study begins where that evaluation stops. We measure selection as its own intervention, comparing full and selected representations within the same readers. The representation targets conversational register rather than general documents. We add a supervised encoder that benefits without selection, and we distill the producer. We also do not treat length as the objective. Normalization is not designed to shorten a call, and its intended effect is on what the reader must infer rather than on how many tokens it reads.

Structured memory for conversational agents is a second neighbor, and it also operates on dialogue. A-MEM turns each interaction into a note with a contextual description, keywords, and tags. It links related notes, updates older notes as new ones arrive, and retrieves the top-k notes by embedding similarity. It is evaluated on LoCoMo, whose conversations average about 9,000 tokens across as many as 35 sessions \citep{xu2025amem}. SimpleMem compresses interactions into compact units at write time, synthesizes related units into more abstract ones, indexes them under several views, and plans retrieval from the inferred intent of each query. It reports a 26.4\% average F1 gain on LoCoMo with up to thirty times fewer inference tokens \citep{liu2026simplemem}.

These systems answer a different question: what should one agent remember about its own continuing relationship with a user. Their memory is mutable and consolidated over time, and access is based on similarity or planned by a language model for each query. Our setting is an analyst's view over millions of completed conversations between other people. Each call's representation is written once and not revised. Access is a deterministic predicate over a fixed vocabulary, so the same slice can be reproduced across a corpus and audited against the source turns, and no model call sits in the selection path.

\subsection{Reducing what the reader must read}

Context compression treats the reader's input length as the cost to reduce, and its main division is whether the compressor sees the question. RECOMP trains extractive and abstractive compressors on the reader's end-task performance. Its compressors can return an empty string when retrieved documents do not help, and they compress to about 6\% of the original length with little loss \citep{xu2024recomp}. AttentionRAG reformulates each query so that its focus falls on a single token, then keeps the context that this token attends to, reaching up to 6.3 times compression \citep{fang2025attentionrag}. Both methods judge relevance with the current question in hand, so the cost of compression recurs with every question.

Question-independent compression avoids that recurrence. LLMLingua-2 distills GPT-4 compression decisions into a token classifier built on a small encoder, and deletes tokens without seeing the task \citep{pan-etal-2024-llmlingua}. One of its results anticipates a pattern in ours. With Mistral-7B as the reader, the compressed MeetingBank prompt scored higher than the original, 76.22 against 66.95 exact match, which the authors attribute to the smaller model's weaker handling of long contexts. ReadAgent splits a long document into pages and compresses each page into a gist before the question is shown. When a question arrives, it looks up the original pages it needs, extending effective context by 3.5 to 20 times on QuALITY, NarrativeQA, and QMSum \citep{pmlr-v235-lee24c}.

These methods establish that weaker readers can benefit from shorter inputs, and that question-independent preparation combines well with question-time lookup. Our selection stage shares that structure but differs in what it operates on and how it decides. Deletion leaves no structure to query, and gists fix in advance which details survive. Our selection operates on a representation that keeps every attributed claim and labels it, so relevance becomes a rule rather than a model judgment. Normalization and selection also appear as separable effects in our results. Compression work does not separate them, because it makes the input clearer and shorter in the same operation.

\subsection{Paying for interpretation once}

Doing interpretive work before questions arrive predates neural models. \citet{fleischman-etal-2003-offline} mined about two million concept--instance relations offline from 15GB of newspaper text, filtered them with a learned classifier, and answered ``Who is'' questions from the resulting repository. Compared with a web-based question-answering system, the repository answered 25\% more questions correctly and did so three orders of magnitude faster. EVAPORATE brings the same logic to language models. It builds structured tables from semi-structured documents, often by having the model write extraction functions instead of reading every document, and reduces the tokens the model must process by a factor of 110 on average across 16 settings of 10k documents each \citep{arora2023evaporate}.

Index-building retrieval systems apply the idea at corpus scale. GraphRAG extracts an entity graph and precomputes summaries of graph communities, independently of any query, in order to answer global questions over corpora of about a million tokens \citep{edge2024graphrag}. RAPTOR recursively clusters and summarizes text into a tree, and raises the best reported accuracy on QuALITY by 20 absolute points when paired with GPT-4 \citep{sarthi2024raptor}. Within conversation analytics, the most natural question-independent representation is a summary, and TWEETSUMM provides about 6,500 human-written extractive and abstractive summaries of customer-care dialogues \citep{feigenblat2021tweetsumm}.

All of these approaches pay once, and each fixes in advance what the prepared representation keeps. Fleischman's repository keeps one relation type. EVAPORATE's tables keep the attributes that were extracted. Summaries and community reports keep what the summarizer judged salient, and GraphRAG and RAPTOR still call a large model at query time. Our representation fixes the unit and the tag vocabulary, but not the set of facts. Every claim remains in the stored call as natural language, so a question nobody anticipated can still be answered from it, and the query path contains only a rule and a small encoder. A summary remains the most important untested alternative in our setting, and Section 7 lists it among the required baselines.

\subsection{Small models for the recurring cost}

The last commitment is economic: the recurring stages should run on small models. Sequence-level knowledge distillation trains a student on the teacher's generated outputs rather than on reference outputs \citep{kim-rush-2016-sequence}, and our normalizer student is trained this way. Distilling step-by-step adds teacher rationales as extra supervision, so that a 770M T5 model outperforms few-shot 540B PaLM on one benchmark using 80\% of the training data \citep{hsieh2023distilling}. FrugalGPT reduces cost at query time instead, by sending each query through a learned cascade of models and stopping once an answer is reliable, which matches GPT-4 with up to 98\% cost reduction \citep{chen2023frugalgpt}. Ask Me Anything shows that reformatting inputs can close much of the gap between model sizes. The model rewrites each input into several open-ended question-answer prompts, the answers are aggregated, and GPT-J-6B matches or exceeds few-shot GPT-3-175B on most benchmarks tested \citep{arora2023ama}.

In each case the saving attaches to a single task or a single query. A distilled task model serves the task it was trained for, a cascade still makes at least one model call per question, and Ask Me Anything asks the reader to reformat its own input every time. Our pipeline distills the one component that all questions share. A single distillation serves every later question, and the reformatting that Ask Me Anything performs inside the reader happens once, upstream, in a different model. A weak reader therefore receives structured input without having to produce it.

\subsection{What is and is not new}

Most individual ingredients have precedent: self-contained units, dialogue rewriting, dialogue-act tags, question-independent preparation, distilled producers, and small readers that benefit from denser input. We do not claim any of them. The contribution lies in their combination for conversation analytics and in its measurement. A conversation is normalized once into attributed, qualified, source-linked statements. Controlled tags then turn evidence selection into a rule rather than a model call. We measure normalization and selection separately, across a supervised encoder and nine prompted readers, against human labels. Finally, a distilled 0.6B producer is substituted under a fixed classifier, which removes large-model calls from the serving path.

\section{Representation and supplementary analyses}

\subsection{Statement structure and programmatic slices}

Example 1 separates three subjects within customer turn 92: deferring the phone line, moving home, and requesting TV setup. The canonical brand tokens expose explicit mentions, while subject tags also retrieve responses that do not repeat a brand. The final statement preserves both future intent and its condition, ``After I move.''

Source references and statement order remain useful. References to ``the phone line'' depend on the earlier offer, and 88.2 retains two related claims about the brands and customer relationship. Self-contained statements are a design objective; source fidelity and preserved qualifications take precedence over forcing every output into an isolated sentence.

The following predicates select statements from Example 1 in the Introduction. \texttt{contains} denotes a literal match against statement text; \texttt{act}, \texttt{obj}, and \texttt{mod} refer to the appended tags. Results retain source order.

\textbf{Table B1. Exact selections from the running example.}

\begin{center}
\small
\setlength{\tabcolsep}{3pt}
\begin{tabular}{p{0.254\linewidth}p{0.437\linewidth}p{0.261\linewidth}}
\toprule
\textbf{Evidence slice} & \textbf{Predicate} & \textbf{Selected statement IDs} \\
\midrule
Explicit Boost Mobile mentions & \texttt{contains("[BOOST\_MOBILE]")} & 88.1, 88.2 \\
Explicit mentions of both brands & \texttt{contains("[BOOST\_MOBILE]") AND contains("[DISH]")} & 88.2 \\
Wireless discussion & \texttt{obj = WIRELESS} & 88.1, 88.2, 89.1, 90.1, 91.1, 92.1, 93.1 \\
Agent's wireless offers & \texttt{speaker = A AND act = offer AND obj = WIRELESS} & 88.1, 89.1 \\
Customer's wireless deferral & \texttt{speaker = C AND act = refuse AND obj = WIRELESS} & 92.1 \\
Customer's intended wireless acceptance & \texttt{speaker = C AND act = accept AND obj = WIRELESS AND mod = intent} & 93.1 \\
Customer's wireless stance (Figure 1) & \texttt{speaker = C AND obj = WIRELESS AND (act = refuse OR act = accept)} & 92.1, 93.1 \\
Move and TV setup & \texttt{obj = MOVE OR obj = TECH\_VISIT} & 92.2, 92.3 \\
\bottomrule
\end{tabular}
\end{center}

These slices illustrate access to the same stored representation. They are not separate end-to-end task evaluations. A task concerning the customer's stance may need both 92.1 and 93.1; a narrow predicate is useful only when it retains the evidence required by the decision.

\subsection{Encoder precision at the reported recall settings}

\textbf{Table B2. Normalized-minus-raw precision on 16,050 held-out calls.} Both input representations fit the context cap; reference labels are teacher-generated.

\begin{center}
\small
\setlength{\tabcolsep}{3pt}
\begin{tabular}{ll}
\toprule
\textbf{Recall setting} & \textbf{Precision difference} \\
\midrule
0.60 & +0.098 \\
0.70 & +0.119 \\
0.80 & +0.156 \\
0.90 & +0.158 \\
\bottomrule
\end{tabular}
\end{center}

The normalized arm leads at all four reported settings. Because the teacher supplies both the normalized text and the reference labels, this comparison measures improved teacher agreement. It does not replace the independent human evaluation in Table 1 or isolate rewriting from deterministic entity canonicalization.

\subsection{Exploratory reader association}

Using the rounded values in Table 2, the Spearman correlation between raw F1 and selected-input F1 is 0.4167. The correlation between raw F1 and the change from raw to selected input is -0.8000.

The second statistic is mathematically coupled: the raw score is also subtracted when computing the gain. For Pearson covariance, the corresponding identity is

\[
\operatorname{Cov}(X, Y - X) = \operatorname{Cov}(X, Y) - \operatorname{Var}(X).
\]

Although this identity does not calculate Spearman correlation, it illustrates why variation and measurement error in the baseline can contribute to an inverse association. The result is therefore descriptive rather than evidence identifying a capability-substitution mechanism.

An exploratory split into four lower- and four higher-raw-F1 readers, excluding the median reader, yields group medians of 0.45305 and 0.77840 on raw input, and 0.72945 and 0.76975 on selected input. The gap between those group medians decreases from 0.32535 to 0.04030. This is distinct from the range across all nine readers, which decreases from 0.4425 to 0.2621.

\subsection{Smaller training-family results}

A separate experiment family based on 7,305 calls produced normalized-input and selected-input classifiers whose performance on the human pilot was near chance. The paired interval for that representation comparison included zero. A student-versus-teacher comparison in the same experiment family favored student input, with a reported interval excluding zero, but involved only a few calls and did not resolve the underlying poor absolute performance.

These outcomes provide no basis for attributing the problem solely to the evaluation set. They may reflect training, representation, calibration, sampling, or other differences in that experiment family. We therefore keep them separate from the production-classifier result in Table 3 and do not use them to establish student superiority.

\subsection{A synthetic call, worked end to end}

\textbf{Example 2.} The exchange below is synthetic, not drawn from the corpus. It shows one bounded window carried through the whole construction: the raw recognizer transcript with its global turn indices, then the normalized, tagged statements. Every transformation named in Section 3.1 appears --- the greeting at turn 1 is dropped, disfluencies and false starts are removed, references are resolved, brands are canonicalized, the redaction token is copied verbatim, and each speaker's stance is kept.

\begin{itemize}

\item \textbf{[1] A:} Hi, thanks for holding, this is Nicole.

\item \textbf{[2] C:} uh yeah, hi, this is Denise Holloway.

\item \textbf{[5] C:} I wanted to add the, uh, the sports package, for the college football this weekend.

\item \textbf{[10] A:} sure, so it's uh \$13.00 a month, but I can do 50\% off, so that's \$6.50.

\item \textbf{[11] C:} OK, yeah, that, that sounds great, let's do it.

\item \textbf{[15] A:} and will you still be using the card ending in [CREDIT\_DEBIT\_NUMBER] for autopay?

\item \textbf{[18] A:} and, oh, are you familiar with our cell phone service, Boost Mobile?

\item \textbf{[19] C:} no, I'm good for now.

\end{itemize}

Dropping the greeting at turn 1, normalization yields:

\begin{quote}\small
\begin{minipage}{\linewidth}\raggedright\setlength{\parindent}{0pt}
\textbf{2.1 C} My name is [DENISE\_HOLLOWAY].\enspace\allowbreak\mbox{\tagpill{tagInform}{inform}\,\tagpill{tagOther}{IDENTITY}\,\tagpill{tagActual}{actual}}\par\smallskip
\textbf{5.1 C} I want to add the sports package for the college football games this weekend.\enspace\allowbreak\mbox{\tagpill{tagRequest}{request}\,\tagpill{tagOther}{SPORTS\_PACKAGE}\,\tagpill{tagActual}{actual}}\par\smallskip
\textbf{10.1 A} The [MULTI\_SPORT\_PACK] is \$13.00 a month, but I can add it at 50\% off, which is \$6.50.\enspace\allowbreak\mbox{\tagpill{tagOffer}{offer}\,\tagpill{tagOther}{SPORTS\_PACKAGE}\,\tagpill{tagActual}{actual}}\par\smallskip
\textbf{11.1 C} Please add the [MULTI\_SPORT\_PACK].\enspace\allowbreak\mbox{\tagpill{tagAccept}{accept}\,\tagpill{tagOther}{SPORTS\_PACKAGE}\,\tagpill{tagActual}{actual}}\par\smallskip
\textbf{15.1 A} Will you still be using the card ending in [CREDIT\_DEBIT\_NUMBER] for autopay?\enspace\allowbreak\mbox{\tagpill{tagAsk}{ask}\,\tagpill{tagOther}{PAYMENT\_METHOD}\,\tagpill{tagActual}{actual}}\par\smallskip
\textbf{18.1 A} Are you familiar with our cell phone service, [BOOST\_MOBILE]?\enspace\allowbreak\mbox{\tagpill{tagOffer}{offer}\,\tagpill{tagWireless}{WIRELESS}\,\tagpill{tagActual}{actual}}\par\smallskip
\textbf{19.1 C} I am not interested in [BOOST\_MOBILE] right now.\enspace\allowbreak\mbox{\tagpill{tagRefuse}{refuse}\,\tagpill{tagWireless}{WIRELESS}\,\tagpill{tagActual}{actual}}
\end{minipage}
\end{quote}

The generic ``sports package'' at turn 5 names no product and is left untagged for entity, while the agent's specific offer at turn 10 canonicalizes to \texttt{[MULTI\_SPORT\_PACK]}; ``let's do it'' at turn 11 is read as an acceptance only because turn 10 established the offer, and the customer's closing ``no, I'm good for now'' keeps its force as a refusal rather than softening into a deferral.

\section{Cost projections and accounting assumptions}

\subsection{Component rates and full projection}

The main comparison uses the strongest prompted configuration, Haiku on full teacher-normalized input. \(B_T\) is its teacher-normalization cost per million calls; it is left explicit because the supplied rates do not resolve that stage. The student pipeline's \$264.85 construction rate is paired only with the student-input classifier's F1 of 0.8387, not the teacher-input classifier's 0.8485. Haiku's quality on student input has not been measured.

\textbf{Table C1. Expanded quality and cost comparison.} Costs are dollars per million calls, rounded; \(n > 1\) projects separate task passes. Add \(B_T\) to every Haiku amount.

\begin{center}
\small
\setlength{\tabcolsep}{3pt}
\begin{tabular}{p{0.609\linewidth}p{0.071\linewidth}p{0.051\linewidth}p{0.051\linewidth}p{0.061\linewidth}p{0.061\linewidth}}
\toprule
\textbf{Serving configuration} & \textbf{STOP F1} & \textbf{n = 1} & \textbf{n = 5} & \textbf{n = 20} & \textbf{n = 50} \\
\midrule
0.6B normalizer + encoder, including normalization & 0.8387 & 272 & 298 & 398 & 599 \\
Haiku 4.5 on full teacher-normalized input, reader cost only & 0.8571 & 1,916 & 9,580 & 38,319 & 95,798 \\
\bottomrule
\end{tabular}
\end{center}

Before rounding, student totals are \$271.53, \$298.25, \$398.45, and \$598.85. Haiku reader-only totals are \$1,915.95, \$9,579.75, \$38,319.00, and \$95,797.50. Ratios in Section 6 use these amounts. The reported marginal encoder pass costs approximately 1/287 of the Haiku pass.

\subsection{Lower-priced prompted baselines}

The original Nova cost projections remain useful context for the broader quality--cost trade-off. Their lower F1 scores make them less informative reference points for the main comparison in Section 6.

\textbf{Table C2. Conditional costs for lower-priced raw-input readers.} Costs are dollars per million conversations, rounded to the nearest dollar. F1 is measured only on STOP; \(n > 1\) projects separately scored tasks at the stated rates. The student row includes its assumed normalization cost.

\begin{center}
\small
\setlength{\tabcolsep}{3pt}
\begin{tabular}{llllll}
\toprule
\textbf{Configuration and cost formula} & \textbf{STOP F1} & \textbf{n = 1} & \textbf{n = 5} & \textbf{n = 20} & \textbf{n = 50} \\
\midrule
0.6B normalizer + encoder: 264.85 + 6.68n & 0.8387 & 272 & 298 & 398 & 599 \\
nova-micro on raw input: 82.41n & 0.4516 & 82 & 412 & 1,648 & 4,121 \\
nova-lite on raw input: 142.21n & 0.6222 & 142 & 711 & 2,844 & 7,111 \\
\bottomrule
\end{tabular}
\end{center}

Under these rates, the student pipeline becomes cheaper than raw-input nova-micro at the fourth task. This crossover concerns alternatives with substantially different observed task quality. The main comparison instead uses Haiku's strongest evaluated input configuration.

\subsection{What the projection includes}

For a fixed corpus, the general model \(B + \sum_q c_q\) allows different selectors, input lengths, and readers to have different marginal costs. The tabulated scenarios hold a per-task rate fixed to illustrate amortization. They estimate serving and exclude development costs such as labels, training, calibration, and schema design.

The normalization and encoder rates must correspond to the evaluated student configuration, including its longer output and any required metadata-generation stage. The reported batch/list-price estimates require reconciliation with hardware or provider, throughput, numerical precision, batch size and concurrency, utilization, token distributions, price date, and discounts. The stated totals are conditional until those inputs are resolved against the run and pricing records.

Alternative serving strategies can change the comparison. Multiple questions can share a prompt; caching can reduce repeated input processing; raw-transcript encoders can share computation across heads. Summaries and retrieval indexes provide other reusable representations. Their costs are informative only alongside quality on the same tasks.

Caching one pooled whole-call encoder vector is also a different computation from re-encoding a question-specific selected span. The tag-head experiments do not establish equal quality for that cached variant. Multi-task evaluation should therefore measure representation reuse, reader reuse, and per-question setup costs separately.

\section{Zero-shot Laya evaluation}

\subsection{Protocol}

We evaluate the released Laya typed-decisions checkpoint on selected normalized statements from the same 66 human-consensus calls, including 17 STOP positives. Zero-shot here means no additional training, fine-tuning, or labeled STOP demonstrations for this task; the released checkpoint already has its own prior training. We do not evaluate Laya on the larger silver-label corpus.

The runner uses \texttt{max\_len = 1024} and \texttt{head\_max\_len = 256}, with each answer option capped at 48 tokens. The original wording uses approximately 165 head tokens: 77 for instructions and 88 for option criteria. It is a compressed task specification, not the complete approximately 1,000-token rubric used by the generative readers. Follow-up variants enrich, shorten, or rephrase that specification within the budget.

The initial run reports two truncated inputs out of 66, both non-STOP, and no empty selections. All 17 positive calls remain in the denominator. This rules out truncation of positive calls in that run; it does not establish that the selected text retained all necessary evidence or that truncation had no effect on negative-call scores.

Each wording is evaluated once with fixed released weights. This is a wording-sensitivity analysis, not a repeated-run or training-seed study. The follow-up wordings were evaluated after the initial result was known, so the analysis is exploratory. We report every variant rather than presenting the highest-scoring wording as a new independent confirmation.

\subsection{Results and interpretation}

\textbf{Table D1. Laya wording sensitivity on the 66-call human cohort.} Laya precision, recall, and F1 use threshold 0.5. ROC-AUC uses the returned STOP scores. The supervised reference is the full-normalized-input ablation classifier in Table 1, not either production-classifier row in Table 3. The comparison shares labels and calls, but differs in training and input scope. Values retain the precision supplied in the experiment report.

\begin{center}
\small
\setlength{\tabcolsep}{3pt}
\begin{tabular}{p{0.521\linewidth}p{0.169\linewidth}p{0.073\linewidth}p{0.073\linewidth}p{0.085\linewidth}}
\toprule
\textbf{Reader / wording} & \textbf{STOP precision} & \textbf{Recall} & \textbf{F1} & \textbf{ROC-AUC} \\
\midrule
Laya: q1\_original & 0.556 & 0.294 & 0.385 & 0.643 \\
Laya: q2\_detailed, enriched definition & 0.333 & 0.118 & 0.174 & 0.607 \\
Laya: q3\_terse, two sentences & 0.429 & 0.529 & 0.474 & 0.733 \\
Laya: q4\_reworded & 0.323 & 0.588 & 0.417 & 0.678 \\
Task-trained encoder, full normalized input & 0.8235 & 0.8235 & 0.8235 & 0.9616 \\
\bottomrule
\end{tabular}
\end{center}

All four Laya variants underperform the supervised reference in both reported metrics. The detailed definition performs worst in this probe; that observation does not establish that richer task definitions are generally ineffective. F1 changes substantially at the fixed threshold, and ROC-AUC also changes by 0.126 across wordings. The differences therefore cannot be attributed entirely to threshold crossings.

ROC-AUC measures ranking without selecting an operating threshold. A threshold change or monotone score recalibration would not remove the observed ranking gap. We do not infer that the four wordings are statistically equivalent: no paired uncertainty estimate is supplied for their differences. The experiment also does not establish that fine-tuning is necessary or sufficient to close the gap. Model choice, task-definition fidelity, evidence selection, and adaptation remain separate factors for future study.

\subsection{Implications for reusable readers}

A task-trained classifier learns a decision boundary from labels, whereas a zero-shot reader must obtain it from the question and definition alongside the evidence. At hundreds of questions, reusable schema-driven readers could reduce the recurring work of fitting, validating, calibrating, and maintaining separate heads. The present experiment leaves that opportunity open while showing that this checkpoint and these definitions do not recover the STOP classifier's quality.

Further comparison should hold task definitions and evidence inputs fixed, testing raw, full normalized, and selected text where context budgets permit. Zero-shot use, limited adaptation, and task-trained heads should be compared at explicit quality requirements, with per-question setup costs reported alongside inference. Wordings should be developed away from the final evaluation set.

\bibliographystyle{plainnat}
\bibliography{clarify_then_focus}

\end{document}